\documentclass[11pt,a4paper]{article}

\usepackage[margin=1in]{geometry}

\usepackage[utf8]{inputenc}
\usepackage[T1]{fontenc}
\usepackage{hyperref}
\usepackage{url}
\usepackage{booktabs}
\usepackage{amsmath}
\usepackage{amssymb}
\usepackage{graphicx}
\usepackage{float}
\usepackage{microtype}
\usepackage[font=small,skip=4pt]{caption}
\usepackage{array}

\title{Diabetic Retinopathy Grading with CLIP-based Ranking-Aware Adaptation:\\
A Comparative Study on Fundus Images}

\author{
   Sungjun Cho \\
  Hong Kong University of Science and Technology (HKUST) \\
  Hong Kong, China \\
  \texttt{schoaq@connect.ust.hk}
}

\begin{document}

\maketitle

\begin{abstract}
Diabetic retinopathy (DR) is a leading cause of preventable blindness, and automated fundus image grading can play an important role in large-scale screening. In this work, we investigate three CLIP-based approaches for five-class DR severity grading: (1) a zero-shot baseline using prompt engineering, (2) a hybrid FCN-CLIP model augmented with CBAM attention, and (3) a ranking-aware prompting model that encodes the ordinal structure of DR progression. We train and evaluate on a combined dataset of APTOS 2019 and Messidor-2 (n=5,406), addressing class imbalance through resampling and class-specific optimal thresholding. Our experiments show that the ranking-aware model achieves the highest overall accuracy (93.42\%, AUROC 0.9845) and strong recall on clinically critical severe cases, while the hybrid FCN-CLIP model (92.49\%, AUROC 0.99) excels at detecting proliferative DR. Both substantially outperform the zero-shot baseline (55.17\%, AUROC 0.75). We analyze the complementary strengths of each approach and discuss their practical implications for screening contexts.
\end{abstract}

\section{Introduction}

Diabetic retinopathy (DR) is a progressive microvascular complication of diabetes and a leading cause of preventable vision loss worldwide. Early and accurate grading of DR severity---ranging from no DR to proliferative DR---is critical for timely clinical intervention. Manual grading by trained ophthalmologists is expensive, time-consuming, and subject to inter-reader variability, which motivates the development of automated grading systems.

Recent vision-language foundation models, particularly CLIP~\cite{Radford21}, have demonstrated strong zero-shot transfer capabilities by learning aligned image-text representations at scale. This raises a natural question: can CLIP-based approaches be adapted for specialized medical image classification tasks such as DR grading, where inter-class differences can be subtle and class distributions are highly imbalanced?

In this paper, we present a systematic comparison of three strategies that build on CLIP for DR grading. The first is a zero-shot baseline that directly applies CLIP without any domain-specific training. The second is a hybrid model that pairs CLIP's visual backbone with a convolutional decoder and CBAM attention~\cite{Woo18} to enable focused lesion localization. The third is a ranking-aware prompting model that learns prompt embeddings which respect the ordinal severity ordering of DR grades. We combine the APTOS 2019~\cite{Karthick19} and Messidor-2~\cite{Decenciere14} datasets to maximize minority-class representation, and we apply resampling and optimal threshold calibration to further mitigate class imbalance.

Our main contributions are:
\begin{itemize}
    \item A controlled comparison of zero-shot, supervised, and prompt-learning CLIP-based strategies under matched data and evaluation conditions.
    \item An analysis of the complementary strengths of the hybrid FCN-CLIP and ranking-aware models, including their distinct precision-recall trade-offs relevant to clinical deployment.
    \item A practical examination of class imbalance mitigation strategies (resampling + threshold calibration) in the DR grading context.
\end{itemize}

\section{Related Work}

\paragraph{Deep learning for DR grading.}
Convolutional neural network (CNN)-based approaches have dominated automated DR grading. Methods based on ResNet, EfficientNet, and related architectures have achieved high accuracy on benchmark datasets such as APTOS and Messidor, often using techniques like focal loss, oversampling, and test-time augmentation to handle class imbalance.

\paragraph{Vision-language models in medical imaging.}
CLIP~\cite{Radford21} and its variants have been explored for zero-shot medical image classification. Domain-specific adaptations, such as FLAIR~\cite{SilvaRodriguez24} for retinal images, demonstrate that text supervision grounded in clinical knowledge can substantially improve performance over general-purpose CLIP features.

\paragraph{Ordinal classification and ranking-aware learning.}
DR grading is naturally an ordinal problem---grades represent increasing severity. Yu et al.~\cite{Yu24} proposed CLIP-DR, which introduces ranking-aware prompt learning that encodes the ordinal relationship between grades into the CLIP similarity space. Our ranking-aware model is informed by this line of work.

\paragraph{Attention mechanisms for lesion localization.}
CBAM~\cite{Woo18} applies sequential channel and spatial attention, which has proven effective for directing model focus toward clinically relevant regions (microaneurysms, hemorrhages, exudates) in medical images.

\section{Methods}

\subsection{Zero-shot CLIP Baseline}

We evaluate CLIP in a fully zero-shot setting using its ResNet-based image encoder. Each DR grade is described by a manually designed text prompt (e.g., ``a fundus photograph showing mild diabetic retinopathy''). The predicted grade is determined by the highest cosine similarity between the image embedding and each grade's text embedding. We experimented with several prompt templates and selected the best-performing formulation. This baseline measures CLIP's out-of-the-box ability to recognize DR features without any domain-specific adaptation.

\subsection{Hybrid FCN-CLIP with CBAM Attention}

The second approach integrates CLIP's visual features into a supervised classification pipeline. The frozen CLIP image encoder (ResNet-based) extracts feature maps from each input image. These feature maps are processed by a fully convolutional network (FCN) decoder comprising convolutional, batch normalization, and ReLU layers. CBAM attention modules are inserted to apply sequential channel and spatial attention, directing the model's focus toward clinically relevant lesion regions. The attended features are globally pooled and passed to a linear classifier. The model is trained with weighted cross-entropy and focal loss (gamma=2.0) to address class imbalance. Architectural details are provided in Appendix~\ref{app:arch}.

\subsection{Ranking-aware Prompting Model}

The third approach introduces learnable prompt vectors for each DR grade to exploit the ordinal structure of DR severity. The CLIP image encoder produces an image embedding $v$, and a text encoder processes grade-specific prompts to produce text embeddings. An image-text similarity matrix is computed via inner products. During training, prompt embeddings are optimized so that similarity scores respect the ordinal rank order: the ground-truth grade should yield the highest score, and more severe grades should score higher than less severe ones. A ranking loss is combined with cross-entropy loss (weighted by alpha=0.7). Class-specific optimal thresholds are calibrated on the validation set, converting the multi-class problem into five one-vs-rest decisions to improve recall on minority classes. Training details are provided in Appendix~\ref{app:training}.

\subsection{Dataset and Preprocessing}
\label{sec:data}

We combine the APTOS 2019~\cite{Karthick19} and Messidor-2~\cite{Decenciere14} datasets to increase minority-class representation. The merged dataset contains 5,406 images across five DR severity grades (Table~\ref{tab:dataset}). We use a stratified 70/15/15\% train/validation/test split with no patient overlap across splits.

\begin{table}[H]
\caption{Class distribution in the combined dataset (APTOS 2019 + Messidor-2).}
\label{tab:dataset}
\centering
\small
\setlength{\tabcolsep}{5pt}
\begin{tabular}{lcccccc}
\toprule
\textbf{Dataset} & \textbf{No DR} & \textbf{Mild} & \textbf{Moderate} & \textbf{Severe} & \textbf{Proliferative} & \textbf{Total} \\
\midrule
APTOS 2019  & 1805 & 370 & 999  & 193 & 295 & 3662 \\
Messidor-2  & 1017 & 270 & 347  &  75 &  35 & 1744 \\
\midrule
Combined    & 2822 & 640 & 1346 & 268 & 330 & 5406 \\
Percentage  & 52.2\% & 11.8\% & 24.9\% & 5.0\% & 6.1\% & 100\% \\
\bottomrule
\end{tabular}
\end{table}

The training set has a highly imbalanced distribution (No DR: 1957, Mild: 449, Moderate: 961, Severe: 188, Proliferative: 229). We apply a resampling strategy to achieve more balanced target counts (No DR: 445, Mild: 200, Moderate: 200, Severe: 180, Proliferative: 180) through combined undersampling and oversampling. Standard augmentations (horizontal/vertical flips, rotation up to 30°, color jitter, affine transforms) are applied to all classes; minority classes additionally receive stronger augmentation (higher flip probability, wider rotation range, random erasing). Details are provided in Appendix~\ref{app:augmentation}.

\section{Experimental Setup}

Both fine-tuned models share the training configuration in Table~\ref{tab:hyperparams}. All models are evaluated on the same held-out test set. We report per-class precision, recall, and F1-score, overall accuracy, macro-averaged precision and recall, and one-vs-rest AUROC.

\begin{table}[H]
\caption{Training hyperparameters.}
\label{tab:hyperparams}
\centering
\begin{tabular}{lc}
\toprule
\textbf{Hyperparameter} & \textbf{Value} \\
\midrule
Batch size         & 32 \\
Learning rate      & 2e-4 \\
Optimizer          & AdamW \\
Weight decay       & 1e-3 \\
Epochs             & 20 \\
Early stopping     & patience 7 \\
Temperature        & 0.2 \\
Loss weight (alpha)& 0.7 \\
\bottomrule
\end{tabular}
\end{table}

\section{Results}

\subsection{Overall Performance}

Table~\ref{tab:comparison} summarizes per-class and overall metrics for all three models on the held-out test set. The ranking-aware model achieves the highest overall accuracy (93.42\%, AUROC 0.9845), followed by the hybrid FCN-CLIP model (92.49\%, AUROC 0.99), with the zero-shot baseline substantially lower (55.17\%, AUROC 0.75).

\begin{table}[H]
\caption{Per-class and overall metrics for the three DR classification approaches.}
\label{tab:comparison}
\centering
\small
\setlength{\tabcolsep}{4pt}
\begin{tabular}{|l|c|c|c||l|c|c|c|}
\hline
\textbf{Metric} & \textbf{Zero-shot} & \textbf{FCN-CLIP} & \textbf{Ranking} & \textbf{Metric} & \textbf{Zero-shot} & \textbf{FCN-CLIP} & \textbf{Ranking} \\
\hline
\multicolumn{4}{|l||}{\textbf{No DR}} & \multicolumn{4}{l|}{\textbf{Severe DR}} \\
\hline
Precision & 0.82 & 0.97 & \textbf{0.99} & Precision & \textbf{0.85} & 0.65 & 0.52 \\
Recall    & 0.61 & \textbf{0.98} & \textbf{0.99} & Recall    & 0.50 & 0.82 & \textbf{0.83} \\
F1        & 0.70 & 0.97 & \textbf{0.99} & F1        & 0.63 & \textbf{0.73} & 0.64 \\
\hline
\multicolumn{4}{|l||}{\textbf{Mild DR}} & \multicolumn{4}{l|}{\textbf{Proliferative DR}} \\
\hline
Precision & 0.20 & \textbf{0.93} & 0.91 & Precision & \textbf{0.86} & 0.66 & 0.81 \\
Recall    & 0.46 & 0.79 & \textbf{0.88} & Recall    & 0.60 & \textbf{0.74} & 0.50 \\
F1        & 0.28 & 0.86 & \textbf{0.89} & F1        & \textbf{0.71} & 0.70 & 0.62 \\
\hline
\multicolumn{4}{|l||}{\textbf{Moderate DR}} & \multicolumn{4}{l|}{\textbf{Overall}} \\
\hline
Precision & 0.54 & 0.96 & \textbf{0.97} & Accuracy  & 0.55 & 0.92 & \textbf{0.93} \\
Recall    & 0.48 & 0.91 & \textbf{0.97} & Macro Prec & 0.65 & 0.83 & \textbf{0.84} \\
F1        & 0.51 & 0.93 & \textbf{0.97} & Macro Rec  & 0.53 & \textbf{0.85} & 0.83 \\
\hline
\end{tabular}
\end{table}

\subsection{Error Distribution}

Table~\ref{tab:errors} shows the distribution of prediction errors by margin. The zero-shot baseline produces a substantial fraction of off-by-two-or-more errors (10\%), indicating frequent severe misclassifications that would be clinically problematic. Both fine-tuned models drastically reduce these, with the ranking-aware model achieving 80\% exact matches and only 2\% large errors.

\begin{table}[H]
\caption{Error distribution across models (percentage of test predictions).}
\label{tab:errors}
\centering
\begin{tabular}{lccc}
\toprule
\textbf{Model} & \textbf{Exact match} & \textbf{Off-by-one} & \textbf{Off-by-two+} \\
\midrule
Zero-shot CLIP   & 60\% & 30\% & 10\% \\
Hybrid FCN-CLIP  & 75\% & 22\% &  3\% \\
Ranking-aware    & 80\% & 18\% &  2\% \\
\bottomrule
\end{tabular}
\end{table}

\subsection{Attention Map Analysis}

\begin{figure}[H]
\centering
\includegraphics[width=0.78\textwidth]{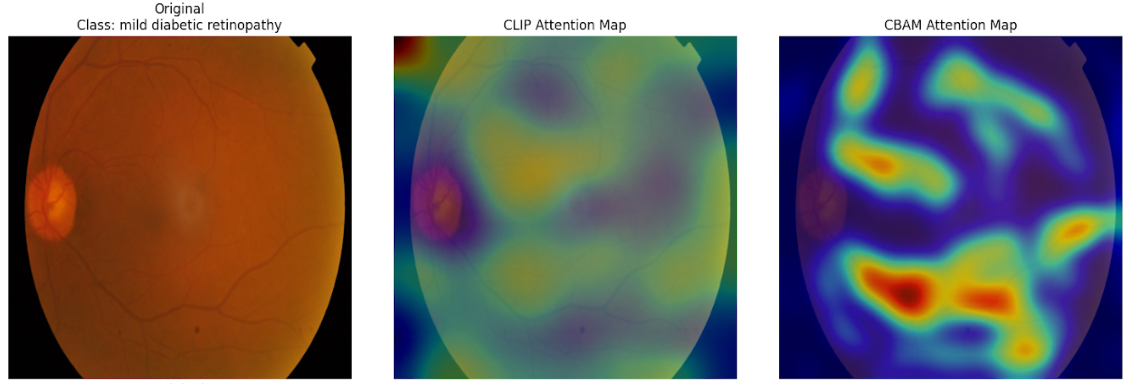}
\caption{Attention map comparison for a mild DR case. Left: zero-shot CLIP (diffuse, anatomy-focused). Right: Hybrid FCN-CLIP with CBAM (concentrated on lesion regions).}
\label{fig:attention}
\end{figure}

Qualitative analysis of attention maps (Figure~\ref{fig:attention}) reveals that CBAM-guided attention in the hybrid model concentrates on clinically relevant microstructures (microaneurysms, hemorrhages, exudates), whereas the zero-shot CLIP baseline shows diffuse attention across coarser anatomical features such as the optic disc and major vessels. This targeted localization explains the hybrid model's advantage in distinguishing adjacent DR grades. Additional attention map visualizations are provided in Appendix~\ref{app:attention}.

\section{Discussion}

\paragraph{Complementary strengths.}
The two fine-tuned models exhibit distinct and complementary strengths. The ranking-aware model achieves the highest overall accuracy and excels at recalling severe DR cases (83\% vs. 82\% for FCN-CLIP), which is especially important in a screening setting where missing severe cases carries high clinical cost. The hybrid FCN-CLIP model, however, substantially outperforms the ranking-aware model on proliferative DR recall (74\% vs. 50\%)---the most critical class for preventing blindness. This suggests that an ensemble of the two approaches could yield even stronger performance, leveraging the ranking-aware model's ordinal sensitivity alongside FCN-CLIP's lesion-detection capabilities.

\paragraph{Precision-recall trade-offs.}
The FCN-CLIP model generally achieves higher precision (fewer false positives), making it more suitable for confirmatory settings. The ranking-aware model prioritizes recall for minority and high-severity classes via threshold calibration, making it more appropriate for population-level screening where sensitivity is paramount.

\paragraph{Imbalance mitigation.}
Class imbalance is a significant practical challenge in DR grading datasets. Our combined resampling and threshold calibration strategy meaningfully improved minority-class performance while preserving overall accuracy. The improvement on Severe DR recall (from 50\% for the zero-shot baseline to 83\% for the ranking-aware model) illustrates the importance of explicit imbalance handling beyond model architecture alone.

\section{Limitations}

This study has several limitations that should be considered when interpreting our results.

\begin{itemize}
    \item \textbf{External validation.} Our evaluation is conducted on a held-out split from the same combined dataset. Performance on independent, out-of-distribution populations (e.g., different imaging devices, patient demographics) remains unknown.
    \item \textbf{Dataset scope.} Although our combined dataset (n=5,406) is reasonably large for DR grading, it represents a limited geographical distribution and may not generalize to all clinical contexts.
    \item \textbf{Clinical deployment gap.} This work demonstrates feasibility in a research setting. Integration into a clinical screening workflow would require prospective validation, calibration studies, and human-in-the-loop evaluation.
    \item \textbf{Computational cost.} The hybrid FCN-CLIP model's memory and compute requirements may limit deployment on resource-constrained hardware, particularly relevant for low-resource settings where automated screening is most needed.
    \item \textbf{Proliferative DR recall.} The ranking-aware model's low recall on proliferative DR (50\%) is a notable limitation given the clinical severity of this class. Future work should directly address this gap.
\end{itemize}

\section{Conclusion}

We presented a systematic comparison of three CLIP-based strategies for diabetic retinopathy grading on fundus images. Both the hybrid FCN-CLIP model and the ranking-aware prompting model substantially outperform the zero-shot CLIP baseline, demonstrating that domain-specific adaptation is essential for effective DR grading. The two fine-tuned models offer complementary strengths: the ranking-aware model provides superior overall accuracy and minority-class recall, while the hybrid FCN-CLIP model excels at detecting proliferative DR through focused lesion localization. These findings suggest that an ensemble or multi-stage approach combining both strategies is a promising direction for future work.

Future directions include integration with more powerful foundation models (e.g., retina-specific vision-language models), semi-supervised or self-supervised adaptation to reduce labeled data requirements, and prospective clinical validation in resource-limited screening settings. We plan to release code and pretrained model weights to support reproducibility.

\section*{Acknowledgments}

We thank the organizers of the APTOS 2019 Kaggle competition and the Messidor-2 dataset for making their data publicly available.


\bibliographystyle{plain}

\appendix
\section{Model Architecture Details}
\label{app:arch}

\paragraph{CLIP backbone.} We use a ResNet-based CLIP image encoder (frozen during training for the hybrid model, updated via prompt learning in the ranking-aware model).

\paragraph{FCN decoder.} The decoder consists of three convolutional blocks (Conv2d $\to$ BatchNorm $\to$ ReLU), with channel widths progressively reducing before global average pooling.

\paragraph{CBAM attention.} Channel attention uses adaptive average pooling followed by a two-layer convolutional bottleneck (reduction ratio 16) with ReLU and sigmoid activations. Spatial attention uses a single 7$\times$7 convolutional layer with sigmoid activation. The implementation is as follows:

\begin{small}
\begin{verbatim}
class CBAM(nn.Module):
    def __init__(self, channels, reduction=16):
        super(CBAM, self).__init__()
        self.channel_gate = nn.Sequential(
            nn.AdaptiveAvgPool2d(1),
            nn.Conv2d(channels, channels // reduction, 1),
            nn.ReLU(),
            nn.Conv2d(channels // reduction, channels, 1),
            nn.Sigmoid()
        )
        self.spatial_gate = nn.Sequential(
            nn.Conv2d(channels, 1, 7, padding=3),
            nn.Sigmoid()
        )

    def forward(self, x):
        x = x * self.channel_gate(x)
        x = x * self.spatial_gate(x)
        return x
\end{verbatim}
\end{small}

\section{Training Details}
\label{app:training}

\begin{figure}[H]
\centering
\includegraphics[width=0.88\textwidth]{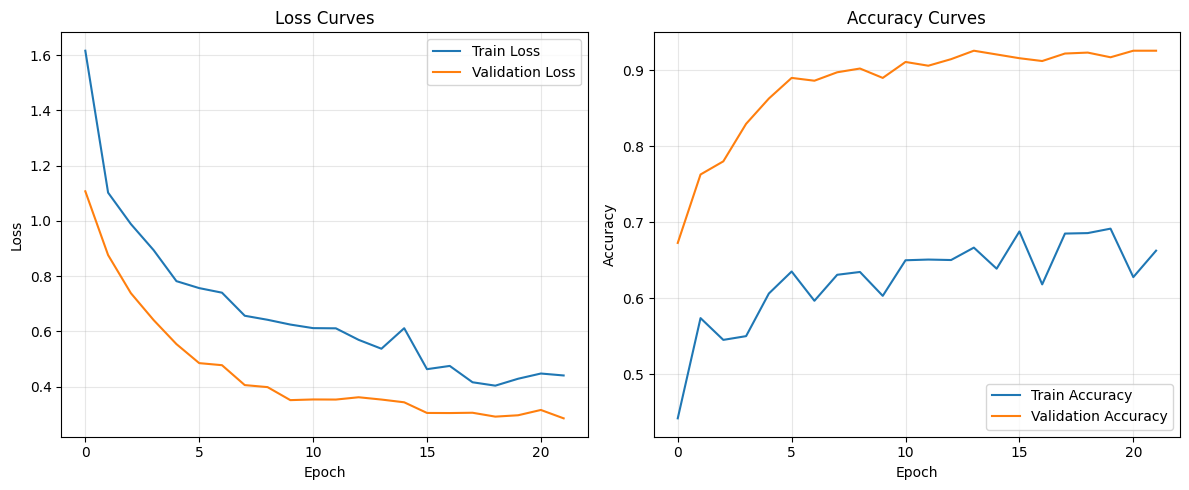}
\caption{Training and validation curves: Ranking-aware prompting model.}
\label{fig:ranking_curves}
\end{figure}

\begin{figure}[H]
\centering
\includegraphics[width=0.88\textwidth]{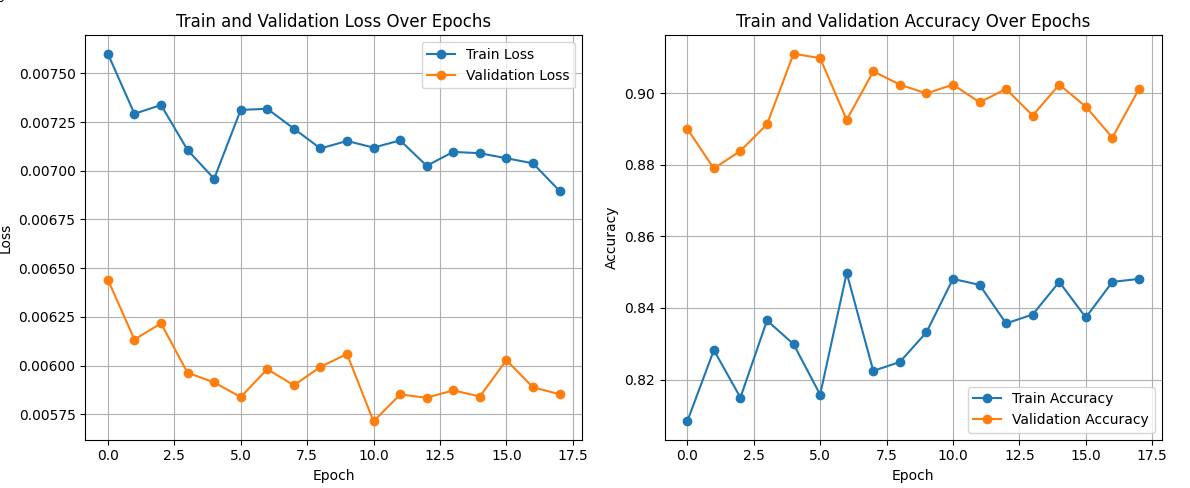}
\caption{Training and validation curves: Hybrid FCN-CLIP model.}
\label{fig:fcn_curves}
\end{figure}

Training and validation accuracy/loss curves for both fine-tuned models are shown in Figures~\ref{fig:ranking_curves} and~\ref{fig:fcn_curves}. Both models converge within 15--20 epochs under early stopping with patience 7.

\section{Data Augmentation and Resampling}
\label{app:augmentation}

\paragraph{Resampling strategy.}
We use a custom oversampling algorithm that samples with replacement from each class to reach target counts, applied only to the training split. The final resampled distribution is shown in Table~\ref{tab:augmentation}.

\begin{table}[H]
\caption{Original vs.\ resampled class distribution in the training set.}
\label{tab:augmentation}
\centering
\begin{tabular}{lccccc}
\toprule
\textbf{Distribution} & \textbf{No DR} & \textbf{Mild} & \textbf{Moderate} & \textbf{Severe} & \textbf{Proliferative} \\
\midrule
Original   & 1957 & 449 & 961 & 188 & 229 \\
Percentage & 51.7\% & 11.9\% & 25.4\% & 5.0\% & 6.0\% \\
\midrule
Resampled  & 445 & 200 & 200 & 180 & 180 \\
Percentage & 37.0\% & 16.6\% & 16.6\% & 15.0\% & 15.0\% \\
\bottomrule
\end{tabular}
\end{table}

\paragraph{Augmentation parameters.}
Standard augmentations applied to all classes: random horizontal/vertical flips (p=0.5), rotation (up to 30°), color jitter (brightness, contrast, saturation), random affine (small translation and scale). Minority class augmentations (Mild, Severe, Proliferative): higher flip probability (p=0.6), wider rotation (up to 35°), stronger color jitter including hue, random erasing (p=0.4).

\section{Additional Visualizations}
\label{app:attention}

\begin{figure}[H]
\centering
\includegraphics[width=0.62\textwidth]{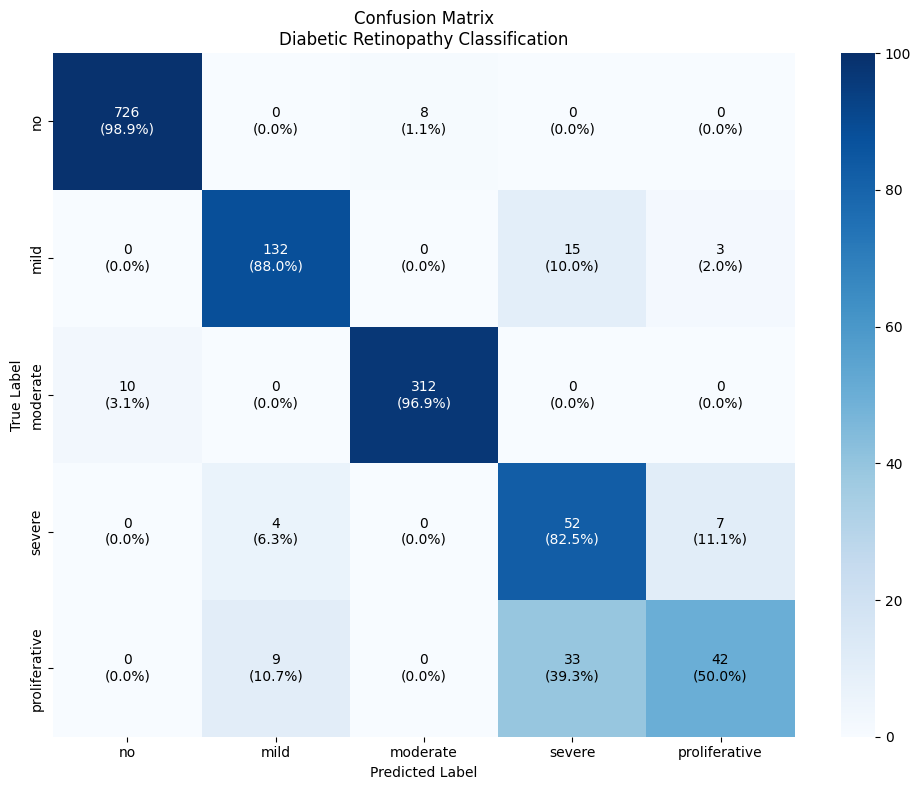}
\caption{Confusion matrix: Ranking-aware prompting model.}
\end{figure}

\begin{figure}[H]
\centering
\includegraphics[width=0.62\textwidth]{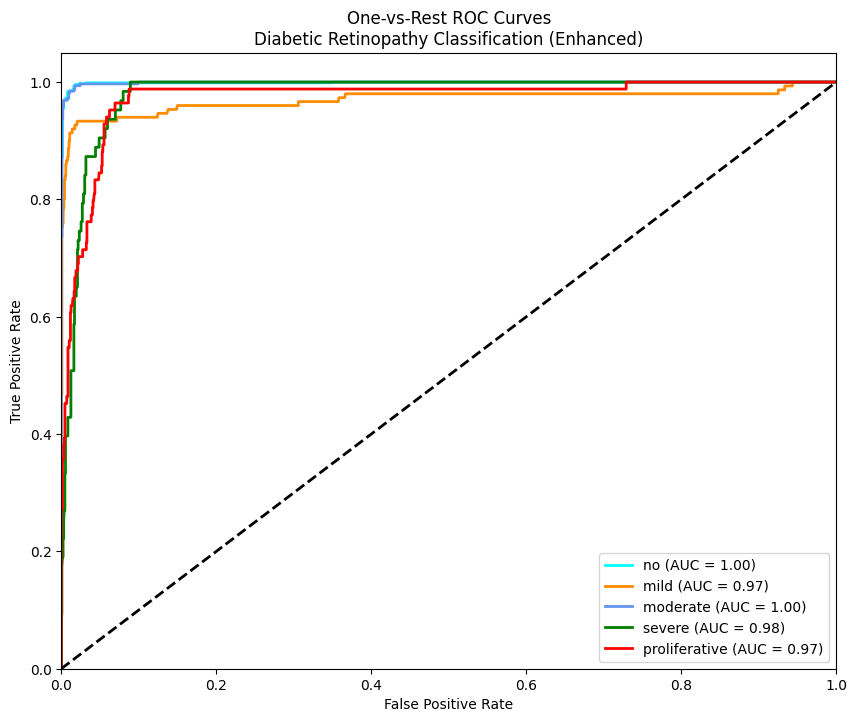}
\caption{One-vs-rest ROC curves: Ranking-aware model.}
\end{figure}

\begin{figure}[H]
\centering
\includegraphics[width=0.62\textwidth]{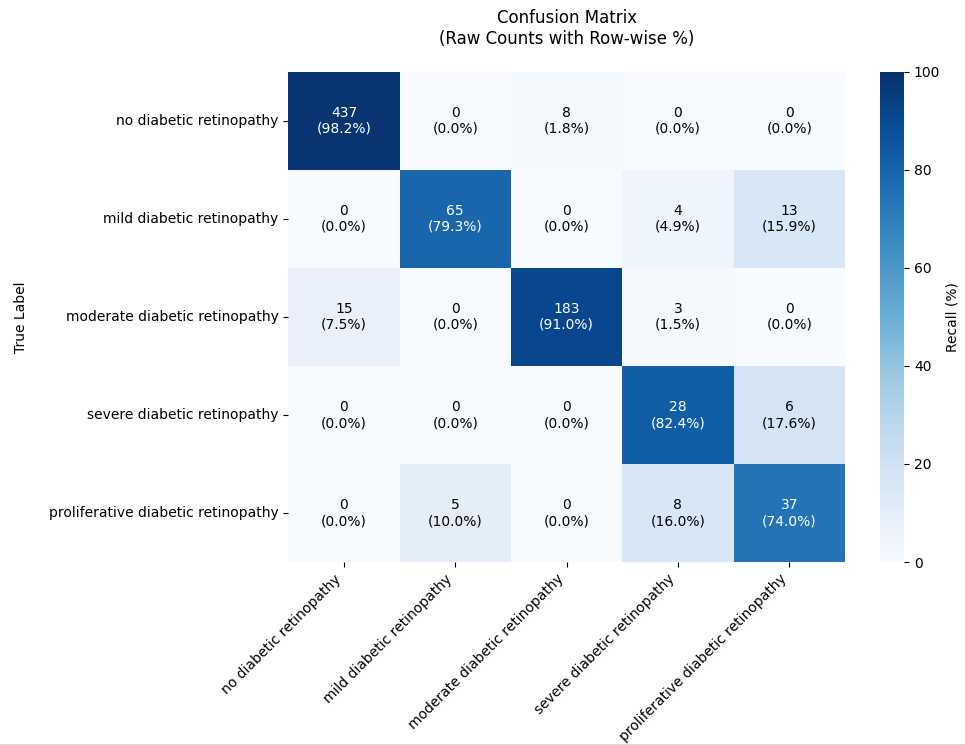}
\caption{Confusion matrix: Hybrid FCN-CLIP model.}
\end{figure}

\begin{figure}[H]
\centering
\includegraphics[width=0.62\textwidth]{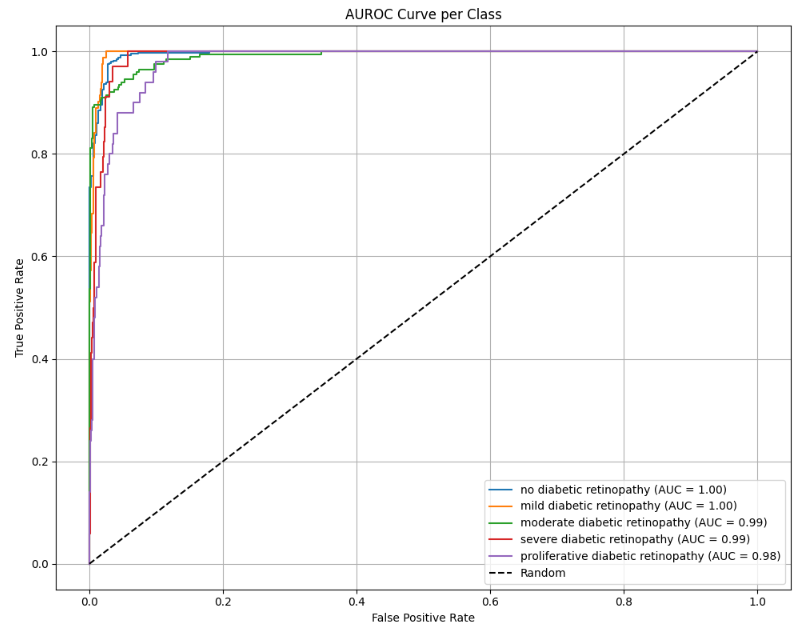}
\caption{AUROC curves per class: Hybrid FCN-CLIP model.}
\end{figure}

\begin{figure}[H]
\centering
\includegraphics[width=0.62\textwidth]{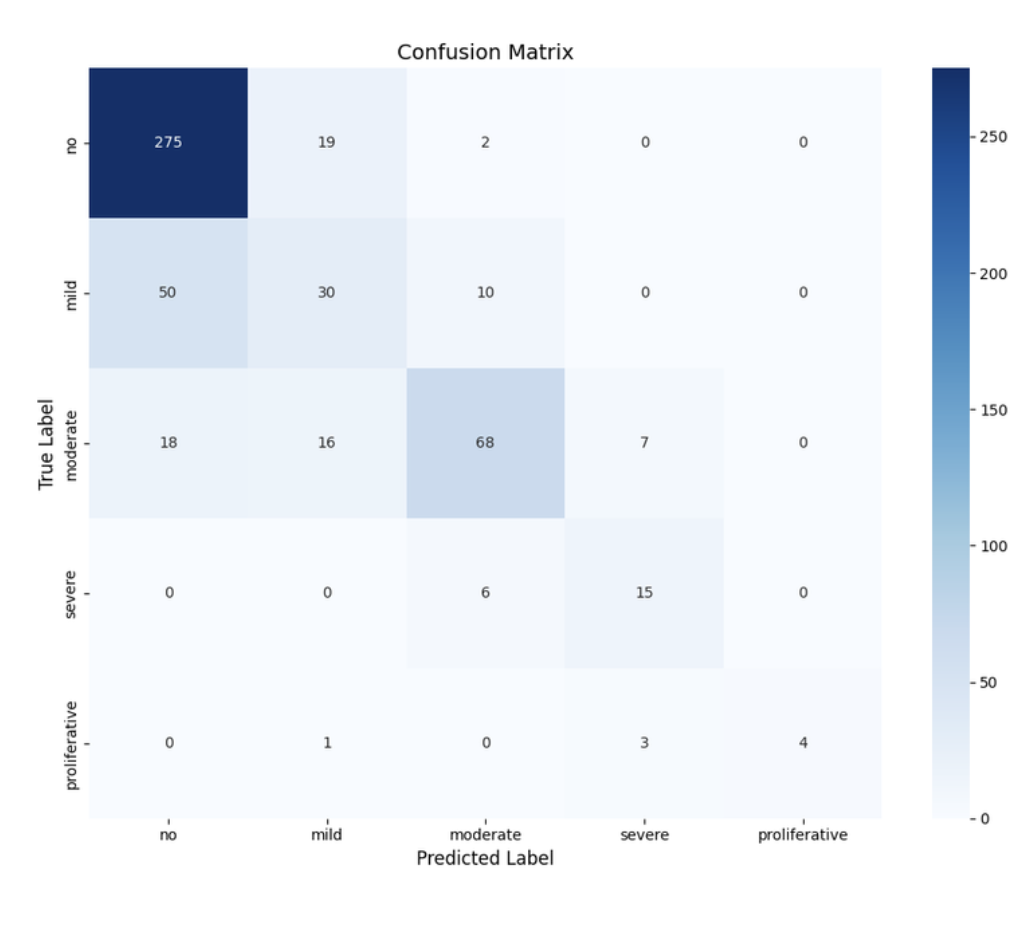}
\caption{Confusion matrix: Zero-shot CLIP baseline.}
\end{figure}

\end{document}